\documentclass[11pt]{article}

\usepackage[preprint]{acl}

\usepackage{times}
\usepackage{latexsym}

\usepackage[T1]{fontenc}
\usepackage[utf8]{inputenc}

\usepackage{microtype}
\usepackage{multirow}

\usepackage{inconsolata}

\usepackage{graphicx}
\usepackage{booktabs}
\usepackage{kotex}
\usepackage{xcolor, soul}

\usepackage{amsmath}
\usepackage{amssymb}
\usepackage{pifont}

\definecolor{myred}{RGB}{243,212,210}
\definecolor{myyellow}{RGB}{254,246,219}
\definecolor{mygreen}{RGB}{218,242,234}
\definecolor{myblue}{RGB}{217,227,241}
\definecolor{mypurple}{RGB}{225,220,235}

\newcommand{\hlyellow}[1]{\sethlcolor{myyellow}\hl{#1}}

\newcommand{\cmark}{\ding{51}}
\newcommand{\xmark}{\ding{55}}

\title{IREA: Intermediate Representation-based Embedding Alignment for Normative RAG}

\author{
  \textbf{Mirae Han\textsuperscript{1}} \quad
  \textbf{Sihyeong Yeom\textsuperscript{2}} \quad
  \textbf{Harksoo Kim\textsuperscript{1}}
  \\
  \textsuperscript{1}Konkuk University,
  \textsuperscript{2}NAVER Cloud
  \\
  \texttt{future26@konkuk.ac.kr} \quad
  \texttt{sihyeong.yeom@navercorp.com} \quad
  \texttt{nlpdrkim@konkuk.ac.kr}
}

\begin{document}

\maketitle

\begin{abstract}
Large language models (LLMs) have shown strong performance across various tasks, but they still struggle with questions involving ethical judgment. Previous studies have attempted to train LLMs on ethical standards, but the diversity and relativity of ethical norms make them difficult to fully internalize in model parameters. As an alternative, we introduce \textit{normative RAG}, a retrieval-augmented approach that supports ethical judgment using external normative knowledge. Normative retrieval involves a distinct asymmetry between context-rich narrative queries and generalized normative statements. Existing factual retrieval methods rely on query-only expansion into a document-like form, making them insufficient for resolving this asymmetry. Therefore, we propose \textbf{I}ntermediate \textbf{R}epresentation-based \textbf{E}mbedding \textbf{A}lignment (\textbf{IREA}), a bidirectional alignment method that maps both text types into a shared \textit{situation-behavior} representation. This representation captures ethically salient contextual and behavioral information in a normalized form, reducing surface-level discrepancies and improving alignment in the embedding space. Experimental results show that IREA improves normative retrieval and downstream ethical judgment across multiple settings, demonstrating the effectiveness of bidirectional alignment for normative RAG.

\end{abstract}

\section{Introduction}

Large language models (LLMs) rely on parametric knowledge acquired during pretraining, which can lead to hallucinations when the required information is not sufficiently represented in their parameters \cite{llm_survey, hallu_survey, park-etal-2025-make}. Retrieval-augmented generation (RAG) mitigates this limitation by leveraging external knowledge to improve response reliability \cite{rag_inital, rag_survey, ic_ralm, lee2026diversifying}. Existing RAG systems have primarily focused on retrieving factual knowledge, which we refer to as \textit{factual RAG}.

\begin{figure}[!t]
    \centering
    \includegraphics[width=\columnwidth]{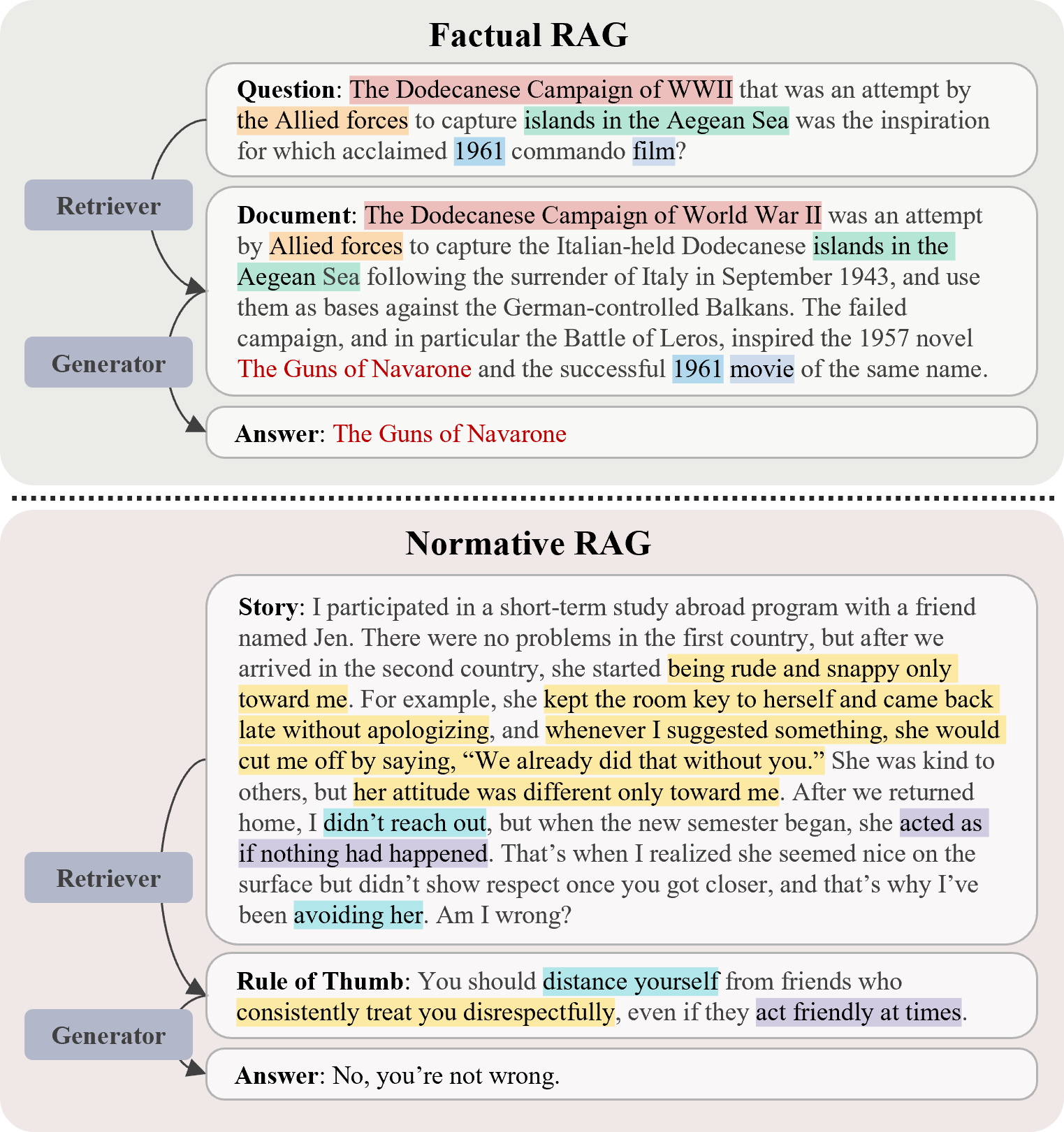}
    \caption{Comparison of factual and normative RAG examples. Same-colored highlights indicate semantic correspondences between texts.}
    \label{fig_1}
\end{figure}

However, LLM limitations extend beyond factuality and also arise in questions requiring ethical judgment \cite{ethical_ambiguity, tao2024cultural}. Previous studies have improved the ethical capabilities of LLMs through fine-tuning on synthetic datasets \cite{culture_park, ConstitutionalAI}. Yet ethical standards are diverse, context-dependent, and evolving, making them difficult to fully internalize in model parameters. Moreover, frequent parameter updates to reflect changing norms may degrade the model's general capabilities \cite{10444954, huang-etal-2025-mitigating, imanov2026mechanistic}.

To address these limitations, we introduce \textit{normative RAG}, which supports ethical judgment by retrieving relevant rules of thumb (RoTs) from external sources. However, as shown in Figure \ref{fig_1}, normative retrieval is more challenging than factual retrieval because queries and RoTs differ substantially in representation. In factual retrieval, queries often share key terms, such as event names or proper nouns, with relevant documents that provide direct evidence for the answer. This makes semantic similarity-based retrieval relatively effective. In contrast, normative retrieval requires matching concrete narrative scenarios to abstract normative principles. In this example, multiple behaviors in the story collectively imply \textit{`rudeness'}, while the corresponding RoT expresses the norm at a more abstract level. Thus, the connection between the story and the RoT is often latent rather than lexical.

Existing factual retrieval methods address query-document mismatch by expanding underspecified queries into document-like representations. Such expansion is effective when additional information improves query-document alignment. Normative retrieval, however, exhibits a distinct asymmetry. The query is already a context-rich narrative, while the RoT is a concise and abstract normative statement. Transforming the query directly into an RoT-like form may over-compress the narrative and discard information needed to identify the underlying moral issue shared by the query and the RoT.

To this end, we propose \textbf{I}ntermediate \textbf{R}epresentation-based \textbf{E}mbedding \textbf{A}lignment (\textbf{IREA}), a bidirectional alignment approach that maps both queries and RoTs into a shared \textit{situation-behavior} representation. In this representation, \textit{situation} specifies the context in which a behavior is interpreted, while \textit{behavior} specifies the action being judged, recommended, or prohibited. Motivated by Frame Analysis \cite{goffman1974frame}, IREA builds on the view that
an action derives its meaning from the contextual frame in which it is interpreted. For example, refusing a request may be appropriate in one context but rude or inappropriate in another. By representing both queries and RoTs in terms of their situation-behavior relations, IREA aligns them at the level of moral meaning while reducing surface-level differences in wording and narrative structure. This structure improves their alignment in the embedding space without discarding information necessary for ethical judgment. Experimental results show that IREA improves RoT retrieval and enhances downstream ethical judgment without additional training. Our main contributions are summarized as follows:

\begin{itemize}

    \item We introduce normative RAG, which extends RAG beyond factual knowledge retrieval to support ethical judgment in LLMs.
    \item We propose IREA, which aligns narrative stories and abstract RoTs through a shared intermediate representation to reduce semantic and structural mismatches in normative retrieval. 
    \item We show that IREA consistently improves RoT retrieval and downstream ethical judgment across benchmark datasets, demonstrating the effectiveness of retrieval-based access to external normative knowledge. 

\end{itemize}

\begin{figure*}[h]
    \centering
    \includegraphics[width=\textwidth]{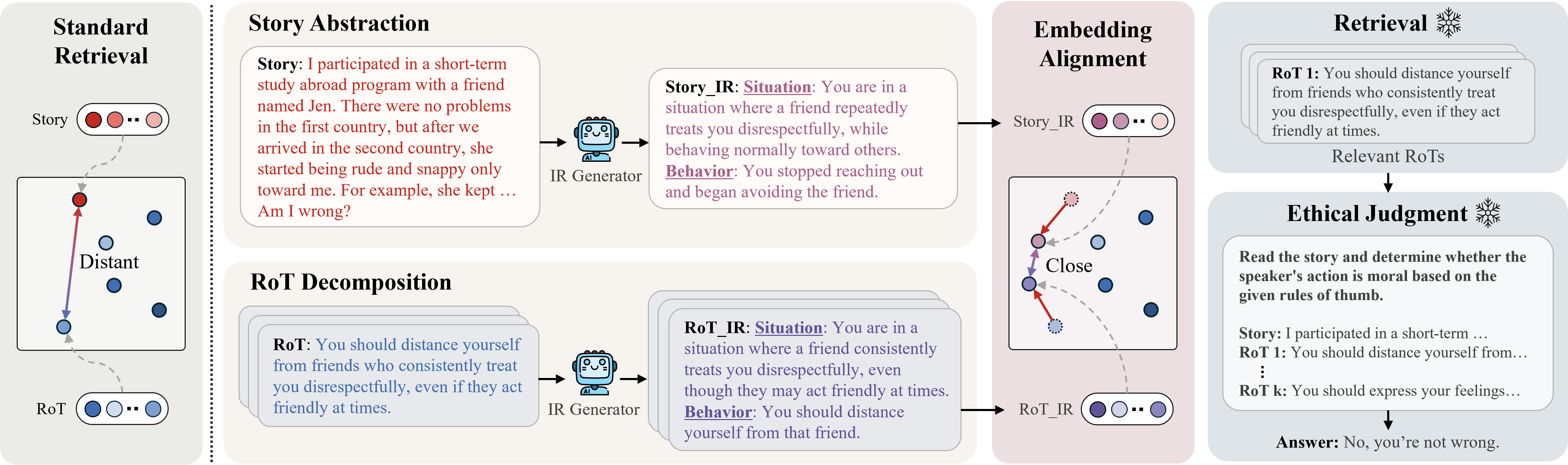}
    \caption{Normative retrieval process of IREA.}
    \label{fig_2}
\end{figure*}

\section{Related Work}
\subsection{LLM Ethics}
Recent LLMs have become more ethically reliable through safety alignment, but they still struggle with complex ethical contexts \cite{fareed2025systematic, hung-etal-2023-walking, hui2025trident, jiashen2025llmscomplicatedethicaldilemma}. They often miss subtle contextual variations, leading to formulaic responses that overlook relevant ethical considerations \cite{soffer2025pitfalls, yu2024greedllamaperformancefinancialvaluealigned}. Moreover, they insufficiently account for cultural diversity and the relativity of ethical standards, resulting in culturally misaligned responses \cite{durmus2023towards, alkhamissi-etal-2024-investigating}. They tend to be biased toward Western-centric values and fail to capture diverse ethical expectations across cultures \cite{shi-etal-2024-culturebank, pham-etal-2024-multi, dwivedi-etal-2025-eticor}.

Previous studies have primarily enhanced the ethical alignment of LLMs through fine-tuning on synthetic datasets \cite{dwivedi-etal-2023-eticor, chan2024enhancingcontentmoderationculturallyaware, jiang2025investigating}. Some studies enrich training data with additional supervision. \citet{10.1007/978-3-031-78172-8_16} incorporated explicit reasoning processes, while \citet{liu-etal-2025-cultural} used role-play data to help models learn cultural norms. Other approaches use ethical principles or human values as reward signals in reinforcement learning \cite{tennant2025moralalignmentllmagents, dai2023safe, chittepu2025reinforcementlearninghumanfeedback}. Training-based methods can improve ethical alignment with a fixed set of norms, but adapting models to diverse ethical contexts through repeated retraining is inefficient. We therefore propose a RAG-based approach that enables norm-grounded ethical responses without additional training.

\subsection{RAG}

RAG has demonstrated strong performance across knowledge-intensive tasks, but its effectiveness depends heavily on the relevance of retrieved documents \cite{cuconasu2024power, EfficientRAG}. To address this limitation, various retriever training methods have been proposed. DPR \cite{dpr}, Contriever \cite{contriever}, and E5 \cite{e5} employ bi-encoder architectures trained to place relevant query-document pairs closer in the embedding space and irrelevant ones farther apart. In addition, some methods jointly train a single retriever across diverse retrieval tasks and objectives, yielding more generalizable embeddings and greater robustness across query types \cite{LLMEmbedder, bge-m3}.

Nevertheless, real-world queries are often incomplete or ambiguous, making it difficult even for strong retrievers to identify relevant documents based solely on the query. To enrich the query with additional context, Query2Doc \cite{query2doc} and HyDE \cite{hyde} generate hypothetical documents for retrieval, while Rewrite-Retrieve-Read \cite{QueryRewriting} reformulates the query into a more retrieval-friendly form. However, these approaches are primarily designed for factual knowledge retrieval and are less effective for abstract expressions such as norms. To address this limitation, we propose a method tailored to normative RAG, where queries and documents differs fundamentally from those in factual RAG.

\section{Methodology}
\subsection{Task Formulation}

Normative RAG aims to retrieve RoTs that are ethically relevant to a given query and use them as supporting evidence for ethical judgment. In this setting, the query is a story that requires ethical assessment, and the documents are RoTs that provide normative guidance. Given a story $s$ and a set of RoTs $\mathcal{R}$, the story and each RoT $r$ are encoded into a shared embedding space using the encoder $f$. The semantic similarity between the resulting vector representations ${v_s}=f(s)$ and ${v_r}= f(r)$ is then computed. The RoTs are ranked according to the similarity scores, and the top-ranked RoTs are retrieved and provided to the LLM for final response generation.
\begin{equation}
    \text{sim}(v_s, v_r) = \frac{v_s \cdot v_r}{\|v_s\|\|v_r\|}, \; r \in \mathcal{R}
\end{equation}

\subsection{IREA}

We propose IREA to mitigate the representational gap between stories and RoTs. Unlike existing factual retrieval approaches that reformulate only the query, IREA maps both stories and RoTs into a shared \textit{situation-behavior} representation. In this schema, \textit{situation} denotes the contextual conditions under which a behavior is ethically interpreted, while \textit{behavior} denotes the action being evaluated in a story or prescribed by an RoT. By representing both text types with the same schema, IREA enables direct comparison between their corresponding situation and behavior components.

IREA is inspired by Goffman's Frame Analysis, which posits that the meaning of an action is constructed within the contextual frame in which it is interpreted, rather than being inherent in the action itself. Accordingly, the same action can carry different social or moral meanings depending on its context. We extend this view to ethical reasoning by conceptualizing ethical judgment as the interpretation and evaluation of behavior within a given situation. This motivates IREA to decompose each text into two components: situation, which captures the context of interpretation, and behavior, which captures the action under moral evaluation.

Specifically, we use an LLM to extract the situation and behavior components from each story and RoT. Given a story $s$, we construct its intermediate representation as $s^{\prime}=(s_\text{sit}, s_\text{beh})$, where $s_{\text{sit}}$ and $s_{\text{beh}}$ represent the extracted situation and behavior, respectively. Similarly, given an RoT $r$, we construct its intermediate representation as $r^{\prime}=(r_\text{sit}, r_\text{beh})$, where $r_{\text{sit}}$ corresponds to the extracted applicability condition and $r_{\text{beh}}$ corresponds to the normative action. Equation \ref{eq_2} formalizes this process, where $g$ denotes the LLM, $i$ the instruction, and $\varepsilon$ the set of in-context examples. The 3-shot examples are constructed from samples disjoint from the test set, and the prompts are provided in Appendix \ref{app_c.3}.
\begin{equation}
\resizebox{0.89\columnwidth}{!}{$
\begin{aligned}
    s^{\prime} &= g(i_s, s, \varepsilon_{s}), \; 
    \varepsilon_s = \{(s_0, s^{\prime}_0), (s_1, s^{\prime}_1), (s_2, s^{\prime}_2)\} \\
    r^{\prime} &= g(i_r, r, \varepsilon_{r}), \; 
    \varepsilon_r = \{(r_0, r^{\prime}_0), (r_1, r^{\prime}_1), (r_2, r^{\prime}_2)\}
\end{aligned}
$}
\label{eq_2}
\end{equation}

Stories often contain concrete events, characters, background details, and emotional expressions that are not always necessary for RoT alignment. IREA therefore filters out incidental details and retains only the situation and behavior directly relevant to ethical judgment. In Figure \ref{fig_2}, the situation is abstracted as \textit{`being in a situation where a friend repeatedly treats you disrespectfully'}, while the behavior is abstracted as \textit{`not reaching out and avoiding the friend'}. This transformation is not a simple summary, but a reconstruction of the story into a situation-behavior intermediate representation at an RoT-comparable level of abstraction.

By contrast, an RoT typically states which behaviors are desirable or undesirable in a given situation, often compressing its applicability condition and its recommended or prohibited behavior into a single sentence. IREA decomposes an RoT into situation and behavior, making these two elements explicit without introducing content absent from the original RoT. In Figure \ref{fig_2}, \textit{`being in a situation where a friend consistently treats you disrespectfully'} corresponds to the situation, while \textit{`distancing yourself from that friend'} corresponds to the behavior. The extracted situation and behavior are concatenated with the tokens \texttt{Situation:} and \texttt{Behavior:} to construct the intermediate representations, as defined in Equation \ref{eq_3}.
\begin{equation}
\begin{aligned}
    s^{\prime} &= [\texttt{Situation:},\, s_{\mathrm{sit}},\, \texttt{Behavior:},\, s_{\mathrm{beh}}] \\
    r^{\prime} &= [\texttt{Situation:},\, r_{\mathrm{sit}},\, \texttt{Behavior:},\, r_{\mathrm{beh}}]
\end{aligned}
\label{eq_3}
\end{equation}

The intermediate representation facilitates story-RoT alignment along two dimensions. First, it provides structural alignment by reconstructing both text types within a shared schema. Although stories and RoTs differ in format, this schema defines corresponding units of comparison, thereby reducing the influence of differences in surface form and discourse structure on similarity computation. Second, it provides semantic alignment across abstraction levels. Stories and RoTs often express the same moral issue at different levels of abstraction. Stories describe concrete events, whereas RoTs present generalized normative statements. Consequently, embeddings computed from the original texts may be affected by details that are irrelevant to ethical judgment. By abstracting stories and decomposing RoTs into situation and behavior components, IREA aligns the two text types using comparable semantic units and abstraction levels.

The generated intermediate representations are encoded into vectors using the same encoder, yielding $v_{s^{\prime}}=f(s^{\prime})$ and $v_{r^{\prime}} = f(r^{\prime})$. We compute the cosine similarity between $v_{s^{\prime}}$ and each $v_{r^{\prime}}$ and retrieve the top-$k$ RoTs with the highest similarity scores. The retrieved RoTs are then provided to the LLM as normative guidance for ethical judgment. Conditioned on these RoTs, the model $h$ generates an ethical response $\hat{y}$ for the given story.
\begin{equation}
\begin{aligned}
\text{sim}(v_{s^{\prime}}, v_{r^{\prime}}) 
    &= \frac{v_{s^\prime} \cdot v_{r^{\prime}}}{\|v_{s^{\prime}}\| \, \|v_{r^{\prime}}\|} \\
R_k(s^{\prime}) 
    &= \operatorname*{arg\, top\text{-}}k\; \text{sim}(v_{s^{\prime}}, v_{r^{\prime}}) \\
\hat{y} 
    &= h(s, R_k(s^{\prime}))
\end{aligned}
\label{eq_4}
\end{equation}

\section{Experiments}
\subsection{Experimental Setups}

\begin{table*}[!t]
\centering
\resizebox{0.9\textwidth}{!}{%
\begin{tabular}{lcccccccccccc}
\toprule
\multirow{2}{*}{Method} & \multicolumn{6}{c}{Scruples}                                                                                                            & \multicolumn{6}{c}{NormAd}                                                                                                              \\ \cmidrule[\arrayrulewidth](r){2-7}  \cmidrule[\arrayrulewidth](l){8-13} 
                        & R@1                  & R@3                  & R@5                  & R@10                 & MRR@10               & nDCG@10              & R@1                  & R@3                  & R@5                  & R@10                 & MRR@10               & nDCG@10              \\ \midrule
\textbf{E5-base-v2}     &                 &                     &                     &                     &                     &                     &                     &                     &                     &                     &                  &                    \\
Naive RAG                    & 21.08                    & 33.77                     & 39.43                    & 49.05                    & 29.23                    & 33.92                    & 21.01                     & 45.16                     & 57.26                     & 70.96                     & 35.94                     & 44.30                     \\
HyDE                    & \underline{29.67}                     & \underline{41.13}                     & \underline{46.32}                     & \underline{54.09}                     & \underline{36.83}                     & \underline{40.93}                  & 22.33                     & 48.35                     & 64.08                   & \underline{79.59}                     & 39.25                     & 48.88                     \\
Query2doc              & 24.56                     & 38.34                     & 44.00                     & 53.07                     & 33.12                     & 37.86                                & \underline{24.81}  & 49.23                  & 61.22                  & 73.54       & 39.62            & 47.75                     \\
Rewrite-Retrieve-Read                 & 24.08                     & 38.34                     & 45.02                     & 53.48                     & 32.79                     & 37.71                & 24.15                    & \underline{50.77}                     & \underline{66.34}                     & \underline{79.59}                     & \underline{40.95}                     & \underline{50.22}                     \\
IREA                    & \textbf{35.47}            & \textbf{52.46}      & \textbf{58.94}         & \textbf{65.51}            & \textbf{43.76}          & \textbf{48.95}            & \textbf{30.31}            & \textbf{64.36}            & \textbf{79.54}            & \textbf{91.09}            & \textbf{49.98}   & \textbf{59.94}            \\ \midrule
\textbf{all-MiniLM-L12-v2}   &                      &                      &                      &                      &                      &                      &                      &                      &                      &                      &                      &                      \\
Naive RAG                  & 15.89                     & 25.72                     & 31.17                     & 39.02                     & 22.43                     & 26.35              & \underline{20.85}                     & \underline{44.17}                     & 55.72                     & 69.75                     & 35.49                     & 43.67                     \\
HyDE                    & \underline{26.26}                     & \underline{40.18}                     & \underline{46.45}                     & \underline{54.77}                     & \underline{34.88}                     & \underline{39.63}            & 18.43                     & 43.73                     & 57.21                  & 70.79           & 33.97                     & 42.77                     \\
Query2doc               & 16.58                     & 26.67                     & 31.99                     & 39.70                     & 23.16                     & 27.08            & 16.89                     & 34.71                     & 44.50                     & 55.94                     & 28.41                     & 34.96                     \\
Rewrite-Retrieve-Read                 & 17.46                     & 28.72                     & 33.22                     & 42.02                     & 24.61                     & 28.73             & 20.74                     & 42.46                     & \underline{57.32}                     & \underline{73.05}                     & \underline{35.50}                     & \underline{44.41}                     \\
IREA                    & \textbf{30.63}          & \textbf{45.77}                 & \textbf{52.80}            & \textbf{61.66}            & \textbf{40.18}            & \textbf{45.32}            & \textbf{25.25}            & \textbf{54.40}            & \textbf{67.93}            & \textbf{80.09}            & \textbf{42.53}            & \textbf{51.58}            \\ \midrule
\textbf{LLM2Vec}   & \multicolumn{1}{l}{} & \multicolumn{1}{l}{} & \multicolumn{1}{l}{} & \multicolumn{1}{l}{} & \multicolumn{1}{l}{} & \multicolumn{1}{l}{} & \multicolumn{1}{l}{} & \multicolumn{1}{l}{} & \multicolumn{1}{l}{} & \multicolumn{1}{l}{} & \multicolumn{1}{l}{} & \multicolumn{1}{l}{} \\
Naive RAG                    & 21.42                     & 34.04                     & 39.97                     & 48.36                     & 29.29                     & 33.82             & \underline{32.84}                      & 59.90                     & 75.30                     & \underline{86.47}                     & 49.56                     & \underline{58.45}                     \\
HyDE                    & \underline{30.70}                     & 42.70                     & 48.70                     & 57.30                     & 38.54                     & 42.99             & 19.69                     & 41.69                     & 56.00                     & 70.02                     & 34.43                     & 42.91                     \\
Query2doc               & 26.88                     & 40.59                     & 47.00                     & 56.07                     & 35.43                     & 40.34            & \textbf{33.39}                     & \underline{61.22}                     & \underline{75.36}                     & 84.38                     & \textbf{49.96}                     & 58.31                     \\
Rewrite-Retrieve-Read                 & 30.22                     & \underline{44.41}                     & \underline{50.48}                     & \underline{60.16}                     & \underline{39.25}                     & \underline{44.22}          & 27.45                     & 55.61                     & 71.73                     & 83.28                     & 44.81                     & 54.08                     \\
IREA                    & \textbf{43.32}                     & \textbf{58.80}                     & \textbf{64.94}            & \textbf{72.92}            & \textbf{52.49}            & \textbf{57.38}            & 31.02$^{\dagger}$            & \textbf{62.60}$^{\dagger}$            & \textbf{78.05}             & \textbf{88.45}            & \underline{49.64}$^{\dagger}$            & \textbf{59.04}$^{\dagger}$            \\ \bottomrule
\end{tabular}}
\caption{\label{tab_1} RoT retrieval performance. The best performance is highlighted in bold, and the second-highest is underlined. Statistical significance is assessed against the strongest baseline using paired bootstrap resampling with 10,000 iterations. All improvements are significant at $p<0.05$ except those marked with $^{\dagger}$.}
\end{table*}

\setlength{\parindent}{0pt}   % 들여쓰기 제거
\setlength{\parskip}{0.7em}   % 문단 간 간격 추가

\textbf{Datasets}
% \subsubsection{Datasets}\label{sec_4.1.1}
We evaluate IREA on the Scruples \cite{lourie2021scruplescorpuscommunityethical} and NormAd \cite{rao-etal-2025-normad} datasets. Scruples consists of moral dilemma scenarios, while NormAd evaluates the ability to apply ethical standards across cultures. As both datasets contain ethical problems that current LLMs struggle to handle, they provide suitable testbeds for our approach. Since Scruples does not include RoTs, we generate an RoT for each sample following \citet{bonagiri2024sageevaluatingmoralconsistency}, using Claude Haiku 4.5 \cite{anthropic2025claudehaiku45} with a temperature of 0.7. This model is used only for RoT construction, reducing potential bias from model-specific generation patterns. For NormAd, we exclude \textit{neutral} samples because they can only be predicted correctly when irrelevant RoTs are provided. Scruples and NormAd comprise 1,466 and 1,818 story-RoT pairs, respectively. The evaluation prompts are detailed in Appendix \ref{app_c.1}.

\textbf{Baselines}
% \subsubsection{Baselines}
We compare IREA against four representative unsupervised approaches: Naive RAG, HyDE \cite{hyde}, Query2doc \cite{query2doc}, and Rewrite-Retrieve-Read (referred to as Rewrite) \cite{QueryRewriting}. Naive RAG retrieves documents based on semantic similarity to the query. HyDE, Query2doc, and Rewrite improve retrieval by reformulating or expanding the original query. HyDE generates a hypothetical document in a zero-shot setting, while Query2doc generates a pseudo-document in a 4-shot setting and concatenates it with the original query to form an expanded query. Rewrite reformulates the query into a retrieval-friendly form using 3-shot examples. We adapt these methods to normative retrieval using the modified prompts described in Appendix \ref{app_c.2}. For a fair comparison, all expanded queries and intermediate representations are generated with GPT-4o-mini-2024-07-18 \cite{achiam2023gpt} at a temperature of 0.7. We use E5-base-v2 \cite{wang2024textembeddingsweaklysupervisedcontrastive}, all-MiniLM-L12-v2 \cite{wang2020minilm}, LLM2Vec \cite{behnamghader2024llm2veclargelanguagemodels} for text embedding. For ethical judgment, we use Qwen-2.5-7B-Instruct \cite{qwen2025qwen25technicalreport}, Gemma-2-9b-it \cite{gemmateam2024gemma2improvingopen} and GPT-4o-mini with temperature set to 0. Based on retrieval performance, we use E5 in all subsequent experiments.

\textbf{Metrics}
We evaluate retrieval performance using Recall@$k$, MRR@10, and nDCG@10. Recall@$k$ measures whether the gold RoT appears within the top-$k$ results, where we set $k \in \{1, 3, 5, 10\}$ considering the database size. MRR@10 evaluates how highly the gold RoT is ranked based on its reciprocal rank. nDCG@10 measures overall ranking quality by discounting relevance scores according to rank. We evaluate ethical judgment performance using Accuracy and F1 score.

\subsection{Normative Retrieval Performance}

\begin{table*}[!t]
\centering
\resizebox{\textwidth}{!}{%
\begin{tabular}{lcccccccccccccccc}
\toprule
\multirow{3}{*}{Method}       & \multicolumn{8}{c}{Scruples}                                                                                                                      & \multicolumn{8}{c}{NormAd}                                                                                                                        \\ \cmidrule[\arrayrulewidth](r){2-9}  \cmidrule[\arrayrulewidth](l){10-17} 
                              & \multicolumn{2}{c}{Top-1}        & \multicolumn{2}{c}{Top-3}        & \multicolumn{2}{c}{Top-5}        & \multicolumn{2}{c}{Top-10}                   & \multicolumn{2}{c}{Top-1}        & \multicolumn{2}{c}{Top-3}        & \multicolumn{2}{c}{Top-5}        & \multicolumn{2}{c}{Top-10}                   \\ \cmidrule[\arrayrulewidth](r){2-9}  \cmidrule[\arrayrulewidth](l){10-17}
                              & Acc.           & F1             & Acc.           & F1             & Acc.           & F1             & Acc.                 & F1                   & Acc.           & F1             & Acc.           & F1             & Acc.           & F1             & Acc.                 & F1                   \\ \midrule
\textbf{Qwen2.5-7B-Instruct} &             &            &             &            &             &            & \multicolumn{1}{l}{} & \multicolumn{1}{l}{} &             &            &             &            &             &            & \multicolumn{1}{l}{} & \multicolumn{1}{l}{} \\
w/o RoT                      & 57.78            & 55.01           & -           & -          & -           & -          & -                    & -                    & 78.82            & 78.10           & -           & -          & -           & -          & -                    & -                    \\
w/ gold RoT                  & 78.99            & 78.39           & -           & -          & -           & -          & -                    & -                    & 95.32            & 95.31           & -           & -          & -           & -          & -                    & -                    \\
Naive RAG                     & 63.78            & 61.95           & 67.19            & 66.06           & 68.14            & 67.54           & 68.76                     & 68.22                 & 90.26            & 90.21           & \underline{90.10}            & \underline{89.99}           & 89.00            & 88.83        & 88.39                     & 88.19                     \\
HyDE                          & \underline{66.51}            & \underline{65.59}           & \underline{68.42}            & \underline{67.77}           & 69.24            & 68.94           & 69.44                     & 69.13                 & 88.50            & 88.49           & 89.82            & 89.76           & \underline{89.82}            & \underline{89.72}        & \underline{89.99}                     & \underline{89.87}                     \\
Query2doc                     & 64.26            & 62.58           & 68.14            & 67.09           & 68.69            & 68.10           & 69.71                     & 69.22                 & \underline{90.98}            & \underline{90.94}           & 89.77            & 89.64           & 88.67            & 88.48        & 88.39                      & 88.20                     \\
Rewrite-Retrieve-Read                       & 63.92            & 62.03           & 67.94            & 66.87           & \underline{70.33}            & \underline{69.76}           & \underline{70.87}                     & \underline{70.47}               & 89.16            & 89.14           & 89.99            & 89.91           & 89.66            & 89.52           & 89.33                     & 89.18                     \\
IREA                          & \textbf{67.87}   & \textbf{66.68}  & \textbf{71.56}   & \textbf{70.88}  & \textbf{71.01}   & \textbf{70.61}  & \textbf{71.90}            & \textbf{71.55}            & \textbf{92.85}   & \textbf{92.82}  & \textbf{92.63}   & \textbf{92.56}  & \textbf{91.36}   & \textbf{91.26}  & \textbf{91.03}            & \textbf{90.93}            \\ \midrule
\textbf{Gemma-2-9b-it}           &             &            &             &            &             &            &                      &                      &             &            &             &            &             &            &                      &                      \\
w/o RoT                      & 63.03            & 60.30           & -           & -          & -           & -          & -                    & -                    & 81.31            & 80.97           & -           & -          & -           & -          & -                    & -                    \\
w/ gold RoT                  & 81.31            & 80.97           & -           & -          & -           & -          & -                    & -                    & 95.43            & 95.42           & -           & -          & -           & -          & -                    & -                    \\
Naive RAG                     & 65.69            & 64.55           & 67.19            & 65.34           & 68.01            & 66.58           & 69.17                     & 67.65                & 90.54            & 90.43           & 91.20            & 91.10           & 91.14            & 91.05           & 90.65                     & 90.54                     \\
HyDE                          & \underline{68.83}            & \underline{67.93}           & 67.94            & 66.33           & 68.42            & 66.86           & 68.42                     & 66.80                     & 90.43            & 90.33           & 91.53            & 91.45           & 91.80            & 91.73           & 92.63                     & 92.58                     \\
Query2doc                     & 66.64            & 65.56           & 68.01            & 66.28           & 68.21            & 66.68           & \underline{69.24}                     & \underline{67.73}                     & \underline{91.86}            & \underline{91.81}           & 91.69            & 91.63           & 91.25            & 91.18           & 91.58                     & 91.51                     \\
Rewrite-Retrieve-Read                       & 66.23            & 65.20           & \underline{68.83}            & \underline{67.03}           & \underline{69.92}            & \underline{68.46}           & 69.17                     & 67.54                     & 90.43            & 90.33           & \underline{92.79}            & \underline{92.73}           & \underline{92.63}            & \underline{92.56}           & \underline{92.96}                     & \underline{92.90}                     \\
IREA              & \textbf{69.71}   & \textbf{69.04}   & \textbf{71.56}  & \textbf{70.34} & \textbf{72.71}  & \textbf{71.62}  & \textbf{70.87}  & \textbf{69.70}            & \textbf{94.55}            & \textbf{94.53}   & \textbf{94.22}  & \textbf{94.19}   & \textbf{93.62}  & \textbf{93.57}   & \textbf{93.62}  & \textbf{93.58}       \\ \midrule
\textbf{GPT-4o-mini}          &             &            &             &            &             &            &                      &                      &             &            &             &            &             &            &                      &                      \\
w/o RoT                      & 70.74            & 70.81           & -           & -          & -           & -          & -                    & -                    & 80.14            & 79.25           & -           & -          & -           & -          & -                    & -                    \\
w/ gold RoT                  &  86.15           & 86.02           & -           & -          & -           & -          & -                    & -                    &  97.85           & 97.85           & -           & -          & -           & -          & -                    & -                    \\
Naive RAG                     & 68.55            & 68.18           & 73.19            & 72.81           & 74.69            & 74.34           & 75.31                     & 74.88                     & 91.03            & 90.95           & 93.95            & 93.92           & \underline{95.21}            & \underline{95.21}           & \underline{94.22}                     & \underline{94.20}                     \\
HyDE                          & \underline{71.08}            & \underline{70.57}           & 73.53            & 73.03           & 74.62            & 74.34           & 74.22                     & 73.63                     & 91.53            & 91.48           & 93.89            & 93.86           & 94.33            & 94.32           & 94.06                     & 94.04                     \\
Query2doc                     & 69.30            & 68.92           & 74.35            & 73.98           & 75.58            & 75.18           & \underline{75.51}                     & \underline{74.97}                     &  \underline{92.30}           & \underline{92.25}           & 93.45            & 93.43           & 93.62            & 93.60           & 93.34                     & 93.33                     \\
Rewrite-Retrieve-Read                       & 69.78            & 69.20           & \underline{74.62}            & \underline{74.16}           & \underline{75.65}             & \underline{75.19}           & \underline{75.51}                     & 74.91                     & 91.75            & 91.70           & \underline{94.61}            & \underline{94.59}           & 94.33            & 94.32           & 93.40                     & 93.38                     \\
IREA                          & \textbf{71.76}   & \textbf{71.51}  & \textbf{76.19}   & \textbf{75.85}  & \textbf{77.97}   & \textbf{77.63}  & \textbf{76.67}            & \textbf{76.23}            & \textbf{95.32}   & \textbf{95.31}  & \textbf{95.60}   & \textbf{95.59}  & \textbf{95.38}   & \textbf{95.37}  & \textbf{95.21}            & \textbf{95.21}            \\ \bottomrule
\end{tabular}}
\caption{\label{tab_2} Ethical judgment performance. `Top-$k$' indicates the setting where the top-$k$ retrieved RoTs are provided to the model. `w/o RoT' denotes the lower-bound setting, in which only the story is provided, whereas `w/ gold RoT' denotes the upper-bound setting, in which the gold RoT is provided. Statistical significance against the strongest baseline is assessed using a paired t-test over five runs with different random seeds ($p<0.05$).}
\end{table*}

\begin{table*}[t]
\centering
\resizebox{0.9\textwidth}{!}{%
\begin{tabular}{lcccccccccccc}
\toprule
\multirow{2}{*}{Method}                                         & \multicolumn{6}{c}{Scruples}                                                                        & \multicolumn{6}{c}{NormAd}                                                                          \\ \cmidrule[\arrayrulewidth](r){2-7}  \cmidrule[\arrayrulewidth](l){8-13} 
                                                                & R@1       & R@3       & R@5       & R@10      & MRR@10    & nDCG@10   & R@1       & R@3       & R@5       & R@10      & MRR@10    & nDCG@10   \\ \midrule
\textbf{IREA}                                                      & \textbf{35.47} & \textbf{52.46} & \textbf{58.94} & \textbf{65.51} & \textbf{43.76} & \textbf{48.95} & \textbf{30.31} & \textbf{64.36} & \textbf{79.54} & \textbf{91.09} & \textbf{49.98} & \textbf{59.94} \\
- Story\_IR                                                        & 19.51          & 32.81       & 38.20          & 47.54    & 27.67          & 32.39          & 24.59          & 51.71          & 65.18          & 78.55          & 41.17          & 50.14          \\
- RoT\_IR                                                          & \underline{31.11}          & \underline{49.05}          & \underline{56.00}          & \underline{65.28}          & \underline{41.68}          & \underline{47.33}          & \underline{27.78}          & \underline{57.76}          & \underline{73.76}          & \underline{85.97}          & \underline{46.07}          & \underline{55.70}          \\
\begin{tabular}[c]{@{}l@{}}- Story\_IR \\ \& - RoT\_IR \end{tabular} & 21.08          & 33.77          & 39.43          & 49.05          & 29.23          & 33.92          & 21.01          & 45.16          & 57.26          & 70.96          & 35.94          & 44.30          \\ \bottomrule
\end{tabular}}
\caption{\label{tab_3} 
The impact of story and RoT intermediate representations.}
\end{table*}

Table \ref{tab_1} shows that IREA consistently outperforms the baselines across experimental settings. Compared with HyDE, the strongest baseline, IREA with E5 increases Recall@10 by 11.42 and 11.50\%p on Scruples and NormAd, respectively. With MiniLM, IREA also yields consistent gains across all metrics on both datasets. With LLM2Vec, IREA achieves the best performance on Scruples, although the improvement over the baselines is smaller on NormAd. Overall, IREA demonstrates robust performance across embedding models and datasets, while baseline performance varies across embedding models. These results suggest that mapping both stories and RoTs into a shared intermediate representation is more effective than query-only reformulation because it aligns them through comparable semantic structures. Appendix \ref{app_a} further shows that IREA remains robust to variations in IR generator capacity, prompt formulation, and the number of in-context demonstrations.

\subsection{Ethical Judgment Performance}

Table \ref{tab_2} shows that IREA consistently outperforms the baselines across all models. Notably, it remains effective with only a few retrieved RoTs and performs closest to the upper bound, where the gold RoT is provided. On Scruples, IREA improves Top-1 Accuracy over the lower bound by 10.09 and 6.68\%p for Qwen and Gemma, respectively, compared with only 1.02\%p for GPT, likely due to its stronger lower-bound performance. This suggests that retrieved RoTs are more beneficial for models with weaker baseline ethical judgment.

Larger gains are observed on NormAd. With IREA, Top-1 Accuracy improves over the lower bound by 14.03, 13.24 and 15.18\%p for Qwen, Gemma, and GPT, respectively. This is likely due to the stronger retrieval performance on NormAd, which provides more relevant RoTs for ethical judgment. Moreover, story-RoT associations are more explicit, allowing retrieved RoTs to be applied with less contextual interpretation. 
Despite slight variation across models in their ability to utilize the retrieved RoTs, these results demonstrate that IREA contributes to enhancing ethical judgment performance. Representative cases are provided in Appendix \ref{app_b.1}.

\subsection{Ablation Study}

We compare IREA with three ablation settings that retain the original form of the story, the RoT, or both. In Table \ref{tab_3}, IREA achieves the highest performance across all metrics, indicating that aligning both text types through a shared intermediate representation is most effective for RoT retrieval. Converting only RoTs considerably degrades performance on both datasets, with the largest drop on Scruples. This suggests that RoT-only conversion may exacerbate representational mismatch by normalizing RoTs while leaving stories in their original narrative form. When only stories are converted, performance also declines, though less substantially. This indicates that abstracting stories into ethically salient components improves retrieval by removing incidental details that can hinder alignment. However, story-only conversion remains insufficient because the compressed form of RoTs limits fine-grained semantic alignment with stories. These results show that transforming both stories and RoTs into intermediate representations is more effective at reducing representational asymmetry and enhancing retrieval performance. Appendix \ref{app_b.2} visualizes this effect in the embedding space.

\subsection{Generalization to an Independent Normative Corpus}

\begin{table}[]
\centering
\resizebox{\columnwidth}{!}{%
\begin{tabular}{lcccccccc}
\toprule
\multirow{2}{*}{Method}      & \multicolumn{2}{c}{Top-1} & \multicolumn{2}{c}{Top-3} & \multicolumn{2}{c}{Top-5} & \multicolumn{2}{c}{Top-10} \\ \cmidrule{2-9} 
                             & Acc.         & F1         & Acc.         & F1         & Acc.         & F1         & Acc.          & F1         \\ \midrule
\textbf{Qwen2.5-7B-Instruct} &          &           &             &           &             &           &              &           \\
w/o RoT                      & \multicolumn{1}{c}{57.78} & \multicolumn{1}{c}{55.01} & \multicolumn{1}{c}{-} & \multicolumn{1}{c}{-} & \multicolumn{1}{c}{-} & \multicolumn{1}{c}{-} & \multicolumn{1}{ c}{-} & \multicolumn{1}{c}{-} \\
Naive RAG                    & 59.41            & 56.85          & 61.46            & 59.57          & 62.55            & 61.25          & 62.89             & 62.02          \\
HyDE                         & \underline{62.55}            & \underline{61.27}          & \underline{62.96}            & \underline{62.21}          & \underline{64.94}            & \underline{64.36}          & \underline{65.48}             & \underline{64.87}          \\
Query2doc                    & 60.03            & 57.52          & 62.35            & 60.46          & 63.44            & 62.29          & 63.17             & 62.24          \\
Rewrite                      & 60.71            & 58.69          & 61.46            & 59.76          & 61.53            & 60.22          & 62.62             & 61.69          \\
IREA                         & \textbf{62.89}            & \textbf{61.32}          & \textbf{64.26}            & \textbf{63.31}          & \textbf{65.42}           & \textbf{64.92}         & \textbf{65.89}             & \textbf{65.47}          \\ \midrule
\textbf{Gemma-2-9b-it}       &          &           &             &           &             &           &              &           \\
w/o RoT                      & \multicolumn{1}{c}{63.03} & \multicolumn{1}{c}{60.30} & \multicolumn{1}{c}{-} & \multicolumn{1}{c}{-} & \multicolumn{1}{c}{-} & \multicolumn{1}{c}{-} & \multicolumn{1}{ c}{-} & \multicolumn{1}{c}{-} \\
Naive RAG                    & 63.03            & 61.63          & 63.37            & 60.50          & 64.39            & 61.86          & 64.19             & 61.82          \\
HyDE                         & \underline{65.69}            & \underline{64.52}          & \underline{65.01}            & \underline{62.86}          & 65.01            & \underline{62.95}          & 64.94             & 62.81          \\
Query2doc                    & 63.44            & 62.11          & 63.98            & 61.15          & 64.87            & 62.44          & 65.01             & 62.45          \\
Rewrite                      & 63.17            & 61.67          & 64.46            & 61.79          & \underline{65.21}            & 62.77          & \underline{66.30}             & \underline{64.00}          \\
IREA                         & \textbf{66.71}            & \textbf{65.75}          & \textbf{66.85}            & \textbf{64.99}          & \textbf{67.53}           & \textbf{65.96}         & \textbf{66.51}             & \textbf{65.05}          \\ \midrule
\textbf{GPT-4o-mini}         &          &           &             &           &             &           &              &           \\
w/o RoT                      & \multicolumn{1}{c}{70.74} & \multicolumn{1}{c}{70.81} & \multicolumn{1}{c}{-} & \multicolumn{1}{c}{-} & \multicolumn{1}{c}{-} & \multicolumn{1}{c}{-} & \multicolumn{1}{ c}{-} & \multicolumn{1}{c}{-} \\
Naive RAG                    & 66.30            & 65.68          & 69.24            & 68.62          & \underline{70.40}            & \underline{69.81}          & \underline{70.60}             & \underline{69.94}          \\
HyDE                         & \textbf{66.78}            & \underline{65.86}          & \underline{70.46}            & \underline{69.93}          & 69.99            & 69.42          & 70.12             & 69.34          \\
Query2doc                    & 66.10            & 65.37          & 68.35            & 67.63          & 70.05            & 69.41          & 69.24             & 68.47          \\
Rewrite                      & 64.53            & 63.73          & 69.51            & 68.79          & 69.99            & 69.26          & 70.53             & 69.75          \\
IREA                         & \underline{66.37}            & \textbf{66.02}          & \textbf{70.60}            & \textbf{70.25}          & \textbf{71.42}           & \textbf{71.12}         & \textbf{72.03}             & \textbf{71.55}          \\ \bottomrule
\end{tabular}}
\caption{\label{tab_4} Ethical judgment performance using RoTs retrieved from a corpus constructed from AITA.}
\end{table}

We evaluate the generalizability of IREA beyond paired settings where each query has a corresponding gold RoT. We construct a corpus of 2,000 RoTs derived from the AITA \cite{bsher2023aitageneratingmoraljudgements} dataset and use Scruples stories as queries. Since the corpus is constructed independently of the queries and lacks query-level relevance annotations, we assess the utility of the retrieved RoTs through downstream ethical judgment performance.

As shown in Table \ref{tab_4}, IREA performs best across diverse settings, indicating that its retrieved RoTs provide useful normative guidance even without query-specific gold RoTs. This setting better reflects practical normative RAG, where queries retrieve normative guidance from an independent knowledge source rather than predefined query-RoT pairs. The consistent gains show that IREA generalizes beyond paired retrieval and helps bridge the representation gap between narrative queries and RoTs through situation-behavior alignment.

\subsection{Robustness to Variations in RoT Quality}

\begin{table}[!t] 
\centering 
\resizebox{\columnwidth}{!}
{\begin{tabular}{lcccccc} 
\toprule Method & R@1 & R@3 & R@5 & R@10 & MRR@10 & nDCG@10 \\ \midrule 
\textbf{Qwen2.5-3B-Instruct} & & & & & & \\ 
Naive RAG & 27.22  & 38.06  & 44.20  & 52.39  & 34.36  & 38.63  \\ 
HyDE & 24.01  & 34.92  & 40.45  & 48.09  & 31.08  & 35.11  \\ 
Query2doc & \underline{30.15}  & \underline{41.81}  & \underline{47.89}  & \underline{56.21}  & \underline{37.63}  & \underline{42.03}  \\ 
Rewrite & 27.90  & 41.47  & 47.48  & 55.80  & 36.40  & 41.03  \\ 
IREA & \textbf{32.88} & \textbf{47.54} & \textbf{53.75} & \textbf{62.55} & \textbf{42.03} & \textbf{46.93} \\ \midrule 

\multicolumn{7}{l}{\textbf{Qwen3-4B-Instruct-2507}} \\
Naive RAG & 24.08  & 34.92  & 41.54  & 51.57  & 31.78  & 36.44\\ 
HyDE & 24.56  & 37.59  & 44.54  & 53.14  & 32.97  & 37.77  \\ 
Query2doc & \underline{27.08}  & 39.43  & 45.63  & 55.46  & 35.21  & 40.00  \\ 
Rewrite & 27.01  & \underline{40.59}  & \underline{47.75}  & \underline{55.73}  & \underline{35.62}  & \underline{40.40}  \\ 
IREA & \textbf{38.95} & \textbf{54.57} & \textbf{61.87} & \textbf{70.53} & \textbf{48.61} & \textbf{53.85} \\ \midrule 

\textbf{Qwen2.5-7B-Instruct} & & & & & & \\ 
Naive RAG & 18.35  & 28.17  & 33.56  & 41.20  & 24.83  & 28.70  \\ 
HyDE & \underline{24.69}  & \underline{36.22}  & \underline{41.81}  & \underline{51.57}  & \underline{32.21}  & \underline{36.77}  \\ 
Query2doc & 20.74  & 31.11  & 36.29  & 45.70  & 27.63  & 31.88  \\ 
Rewrite & 20.33  & 31.58  & 36.43  & 44.41  & 27.20  & 31.28  \\ 
IREA & \textbf{25.24} & \textbf{37.93} & \textbf{43.32} & \textbf{53.00} & \textbf{33.32} & \textbf{37.97} \\ \midrule

\textbf{Qwen2.5-14B-Instruct} & & & & & & \\ 
Naive RAG & 20.60  & 31.79  & 38.47  & 46.38  & 27.95  & 32.32  \\ 
HyDE & \underline{26.47}  & \underline{38.61}  & \underline{43.59}  & \underline{53.00}  & \underline{34.17}  & \underline{38.63}  \\ 
Query2doc & 24.08  & 36.97  & 43.04  & 50.82  & 32.18  & 36.62  \\ 
Rewrite & 22.44  & 34.79  & 40.38  & 49.39  & 30.45  & 34.95  \\ 
IREA & \textbf{33.83} & \textbf{47.61} & \textbf{53.82} & \textbf{63.37} & \textbf{42.58} & \textbf{47.51} \\  
\bottomrule 
\end{tabular}} 
\caption{\label{tab_5} Retrieval performance using RoTs generated by models of different sizes.} 
\end{table}

To evaluate the robustness of IREA to variations in RoT quality, we compared retrieval performance using RoTs generated by models of different sizes. Using generator capacity as a proxy for RoT quality, IREA achieves the best performance across all settings, as shown in Table \ref{tab_5}. Although absolute retrieval performance varies across RoT generators, IREA consistently maintains its advantage over the baselines. In terms of Recall@10, IREA outperforms the strongest baseline by 6.34, 14.80, 1.43, and 10.37\%p for the 3B, 4B, 7B, and 14B settings, respectively. These results suggest that the effectiveness of IREA is not confined to a particular level of RoT quality and remains relatively robust to variations in RoT quality.

\subsection{Error Propagation Analysis}

\begin{table*}[t]
\centering
\resizebox{\textwidth}{!}{
\begin{tabular}{c|p{1.25\textwidth}}
\toprule
\textbf{Error Type} & \multicolumn{1}{c}{\textbf{Example}} \\ 
\midrule

Story\_IR &
\textbf{Story Excerpt:} 
Yesterday we were in my bed watching some Harry Potter movies and she was going through her gallery on the phone and in the corner of my eye i saw what it look like a photo of her with another guy and i asked if i could see the photo... and after years of being in a relationship with her never trusting me, I decided I had enough.

\textbf{Gold RoT:} 
You should end a relationship where your partner consistently refuses to trust you despite your faithful behavior and honesty, rather than continuing to invest in someone who cannot reciprocate basic trust.

\textbf{Story\_IR:}
You are in a long-term romantic relationship where trust issues have escalated, leading to constant accusations and emotional strain between partners, ultimately resulting in the decision to end the relationship. You expressed a desire for your partner to trust you after years of accusations and emotional turmoil, and \hlyellow{you confronted her about a photo that raised your suspicions}, which she reacted to defensively.

\textbf{Retrieved RoT:}
You should respect your partner's boundaries and feelings in a committed relationship, and take responsibility when your actions—like taking intimate photos of an ex—violate the trust and exclusivity you have promised.
\\ \midrule

RoT\_IR & 
\textbf{Story Excerpt:} 
She said she talked to our boss months in advance and told him to keep it a secret. She never asked me if I was okay with using my leave for an overseas holiday, and I told her that I would refuse to go as she should have asked me first.

\textbf{Gold RoT:} 
You should consult your partner about major decisions that affect them both, rather than making unilateral plans that commit their time and resources without their input.

\textbf{RoT\_IR:}
You are in a relationship where you are faced with a significant decision that impacts both you and your partner, \hlyellow{such as moving to a new city, changing jobs, or making financial investments}. You should discuss and consult your partner regarding these major decisions, ensuring that their opinions and preferences are considered before making any commitments that affect both of you.

\textbf{Retrieved RoT:}
You should set boundaries about your labor and time, and you are not obligated to take on projects you don't want to do, even if others disagree with your reasoning.
\\ 

\bottomrule
\end{tabular}}
\caption{\label{tab_6} Examples of IR generation errors that propagate to RoT retrieval failures.}
\end{table*}

Table \ref{tab_6} presents representative retrieval failures caused by IR generation errors. In the first case, the Story\_IR captures the prolonged trust conflict but retains the concrete action of asking to see a photo rather than abstracting the target behavior of ending the relationship due to repeated distrust. As a result, the Story\_IR embedding aligns with norms about trust and photos, leading to retrieval of an incorrect RoT. The second case shows a retrieval failure caused by over-specification in the RoT\_IR. While the gold RoT broadly applies to unilateral decisions affecting a partner’s time or resources, the RoT\_IR adds specific scenarios absent from the original, such as moving or changing jobs. These details introduce semantic signals not present in the story, weakening Story\_IR and RoT\_IR alignment. Overall, these cases show that decomposition alone is insufficient, and effective retrieval requires both representations to maintain a comparable level of abstraction.

\subsection{Effectiveness-Efficiency Trade-off}

\begin{table}[]
\centering
\resizebox{\columnwidth}{!}{%
\begin{tabular}{l|c|c|c|c}
\toprule
\textbf{Method} & \textbf{Cost type} & \textbf{Latency (s) $\downarrow$} & \textbf{Token usage $\downarrow$}  & \textbf{nDCG@10 $\uparrow$} \\ \midrule
HyDE             & Online             & 4.234               & \textbf{605.3}                 &  \underline{40.93}           \\
Query2doc        & Online             & \textbf{1.269}              & 2224.4                 &  37.71   \\
Rewrite          & Online             & 4.547              & \underline{2064.2}                 &  37.86 \\
 \midrule
IREA (Story\_IR) & Online             & \underline{1.974}              & 2290.6                 & \multirow{2}{*}{\textbf{48.95}}   \\ 
IREA (RoT\_IR)   & Offline            & 1.641              & 440.4                  &     \\ \bottomrule
\end{tabular}}
\caption{\label{tab_7} Comparison of retrieval performance and computational efficiency. Results are averaged over all samples in the dataset.}
\end{table}

In Table \ref{tab_7}, IREA achieves the second-lowest latency, reducing it by 53.4 and 56.6\% compared with HyDE and Rewrite, respectively. Although its token usage is the highest, the increase over Rewrite and Query2doc is modest relative to its larger gains in retrieval performance. These results indicate that IREA improves retrieval effectiveness while maintaining competitive efficiency. IREA also converts RoTs into intermediate representations, introducing additional generation cost. However, RoT\_IR generation requires few tokens and low latency because RoTs are typically concise single-sentence statements. Since RoT\_IRs are generated offline and indexed in advance, they add little to online retrieval latency. Even when both intermediate representations are generated online, their combined latency remains lower than that of HyDE and Rewrite. These results indicate that IREA provides a favorable trade-off between retrieval effectiveness and efficiency.

\section{Conclusion}
To address the ethical limitations of LLMs, we introduced normative RAG and proposed IREA, an embedding alignment method for normative retrieval. IREA maps narrative queries and RoTs into a shared situation-behavior representation, reducing representational, semantic, and structural mismatches and enabling more precise embedding-based retrieval. Experimental results show that IREA consistently outperforms existing approaches, highlighting the limitations of factual retrieval methods when applied to normative retrieval. These findings demonstrate the potential of extending RAG beyond factual knowledge retrieval to normative reasoning and more socially responsible LLMs.

% \subsection{References}

% The \LaTeX{} and Bib\TeX{} style files provided roughly follow the American Psychological Association format.
% If your own bib file is named \texttt{custom.bib}, then placing the following before any appendices in your \LaTeX{} file will generate the references section for you:
% \begin{quote}
% \begin{verbatim}
% \bibliography{reference}

% \end{verbatim}
% \end{quote}

\section*{Limitations}

In this paper, we propose an approach that enhances normative retrieval through embedding alignment between narratives and RoTs, and demonstrate its potential to mitigate the ethical limitations of LLMs. Nevertheless, IREA still has limitations and open questions that remain to be addressed in future work.

\begin{itemize}

    \item \textbf{Expressiveness of the Situation-Behavior Schema.} The situation–behavior representation used in IREA is not intended to fully model all ethical problems, but rather to serve as a minimal schema for mapping narratives and RoTs into a shared representation. Information such as intent, consequences, and relationships between actors can be incorporated into the situation when relevant to the ethical interpretation of the behavior. However, more complex ethical structures, such as conflicts or priority relations among multiple norms, may not be sufficiently captured by these two components. Future work could explore more flexible intermediate representations that introduce additional ethical dimensions or dynamically incorporate the elements required by each problem.
    
    \item \textbf{Domain Dependence of IR Generation.}
    The effectiveness of IREA may also depend on how intermediate representations are generated for different ethical domains. Different domains may require slight domain-specific adjustments to the prompt to capture relevant normative information, which may limit the generalizability of a fixed prompt. However, prompt design is not the primary focus of this work. Our focus is instead on using shared intermediate representations to reduce the representational gap between narratives and RoTs. Future work could investigate more adaptive prompting or representation-generation strategies across a broader range of ethical domains.
    
    \item \textbf{Limited Evaluation Setting.} The effectiveness of IREA was demonstrated on two distinct datasets addressing universal ethics and cultural ethics, respectively. However, these datasets are insufficient to represent the diversity of ethical norms. This limitation arises from the scarcity of publicly available datasets that comprehensively cover the wide range of ethical judgment problems that normative retrieval aims to address, resulting in a constrained evaluation setting. Therefore, future work will explore datasets encompassing more diverse norms and ethical judgment scenarios to more systematically assess the generalizability of IREA.

\end{itemize}

% \section*{Acknowledgments}

% This document has been adapted
% by Steven Bethard, Ryan Cotterell and Rui Yan
% from the instructions for earlier ACL and NAACL proceedings, including those for
% ACL 2019 by Douwe Kiela and Ivan Vuli\'{c},
% NAACL 2019 by Stephanie Lukin and Alla Roskovskaya,
% ACL 2018 by Shay Cohen, Kevin Gimpel, and Wei Lu,
% NAACL 2018 by Margaret Mitchell and Stephanie Lukin,
% Bib\TeX{} suggestions for (NA)ACL 2017/2018 from Jason Eisner,
% ACL 2017 by Dan Gildea and Min-Yen Kan,
% NAACL 2017 by Margaret Mitchell,
% ACL 2012 by Maggie Li and Michael White,
% ACL 2010 by Jing-Shin Chang and Philipp Koehn,
% ACL 2008 by Johanna D. Moore, Simone Teufel, James Allan, and Sadaoki Furui,
% ACL 2005 by Hwee Tou Ng and Kemal Oflazer,
% ACL 2002 by Eugene Charniak and Dekang Lin,
% and earlier ACL and EACL formats written by several people, including
% John Chen, Henry S. Thompson and Donald Walker.
% Additional elements were taken from the formatting instructions of the \emph{International Joint Conference on Artificial Intelligence} and the \emph{Conference on Computer Vision and Pattern Recognition}.


\bibliography{reference}

\appendix

% \clearpage

\section{Robustness Analysis}
\label{app_a}

\subsection{Effect of IR Generator Capacity}
\label{app_a.1}

\begin{figure}[!h]
    \centering
    \includegraphics[width=\columnwidth]{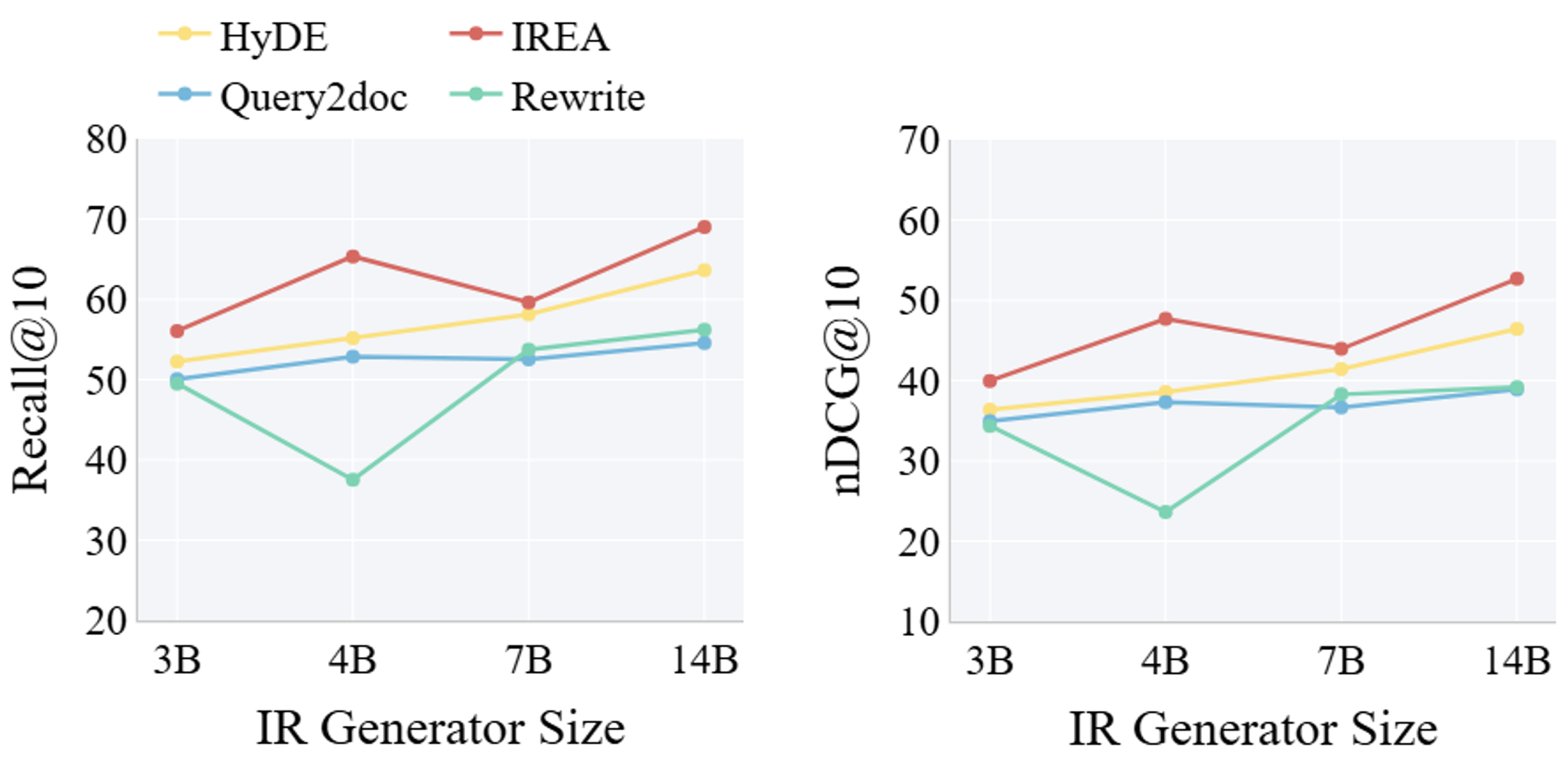}
    \caption{Retrieval performance with different IR generator sizes.}
    \label{fig_3}
\end{figure}

To examine the effect of IR generator capacity on retrieval performance, we vary the generator size within the Qwen model family. Specifically, we use Qwen2.5-3B-Instruct, Qwen3-4B-Instruct-2507, Qwen2.5-7B-Instruct, and Qwen2.5-14B-Instruct as IR generators. Figure \ref{fig_3} presents the results on Scruples. Overall, retrieval performance improves as generator size increases, suggesting that larger generators are more likely to produce retrieval-effective intermediate representations. In particular, IREA consistently achieves the highest Recall@10 and nDCG@10 across all generator sizes. It remains effective even with smaller generators and attains its best performance with the largest generator. Although the baselines also benefit from increased generator capacity, their improvements are relatively modest. These results indicate that IREA is robust across  generator scales and can effectively leverage stronger generators to improve retrieval performance.

\subsection{Prompt Robustness}

\label{app_a.2}

\begin{figure}[t]
    \centering
    \includegraphics[width=\columnwidth]{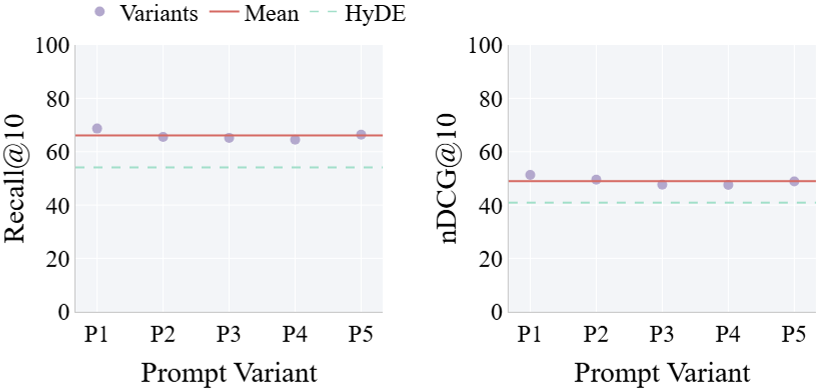}
    \caption{Retrieval performance across prompt variants. `P1' corresponds to the original prompt used in IREA.}
    \label{fig_4}
\end{figure}

To assess the prompt sensitivity of IREA, we construct five paraphrased prompt variants and evaluate retrieval performance on Scruples. As shown in Figure \ref{fig_4}, IREA exhibits minimal variation across prompt variants for both Recall@10 and nDCG@10. For Recall@10, the results remain largely stable, with individual variants closely aligned with the average performance. A similar trend is observed for nDCG@10, indicating that ranking quality is also stable under prompt paraphrasing. Moreover, IREA consistently outperforms HyDE across all variants, suggesting that its retrieval gains are attributable to the introduction of shared intermediate representations rather than to prompt-specific wording. These results demonstrate that IREA is robust to semantically equivalent prompt variations and exhibits limited sensitivity to prompt formulation.

\begin{table}[!t]
\centering
\resizebox{0.85\columnwidth}{!}{%
\begin{tabular}{c|c|c|c}
\toprule
\textbf{Method} & \textbf{Recall@10} & \textbf{MRR@10} & \textbf{nDCG@10} \\ \midrule
0-shot & 64.80                & 41.86               & 47.31                \\
1-shot & \underline{65.55} ±2.37                  & 42.83 ±2.21               & 48.25 ±2.23                \\
3-shot & 65.51 ±2.84                  & \underline{43.76} ±2.03               & \underline{48.95} ±2.22                \\
5-shot & \textbf{66.87} ±3.76                  & \textbf{45.22} ±3.75               & \textbf{50.39} ±3.74                \\ \bottomrule
\end{tabular}}
\caption{\label{tab_8} Retrieval performance with different numbers of in-context demonstrations. Results are reported as the mean and standard deviation across five demonstration sets.}
\end{table}

\begin{table*}[!t]
\centering
\resizebox{\textwidth}{!}{
\begin{tabular}{c|c|p{1.25\textwidth}}
\toprule
\textbf{Retrieval} & \textbf{Judgment} & \multicolumn{1}{c}{\textbf{Example}} \\ 
\midrule

\cmark & \cmark &
\textbf{Story Excerpt:} 
The next thing I know, one of them takes a handful of snow and rubs it on my face. I turn and I punch him 3 times in the head, and on the third hit he falls to the ground.

\textbf{Retrieved RoT:} 
You should not use physical violence in response to playful or minor provocations, even if you have warned someone beforehand; proportional responses and walking away 
are more appropriate.

\textbf{Judgment:} Wrong
\\ \midrule

\xmark & \cmark & 
\textbf{Story Excerpt:} 
But she didn’t mention it and seemed a little protective of her phone which made me suspicious. I didn’t go through her phone but I did mention it this morning and asked that she be more transparent.

\textbf{Retrieved RoT:} 
You should communicate your concerns about relationship boundaries directly and honestly with your partner, rather than avoiding the issue or making assumptions about their feelings.

\textbf{Judgment:} Right
\\ \midrule
\cmark & \xmark & 
\textbf{Story Excerpt:} 
I told her that I want to marry her but I want to wait... To which I responded with what I usually say like it’s too soon, we haven’t even moved in together yet.

\textbf{Retrieved RoT:} 
You should take time to build a strong foundation in a relationship before making major commitments like marriage, rather than rushing into life-changing decisions based on early emotions.

\textbf{Judgment:} Wrong
\\ \midrule
\xmark & \xmark &
\textbf{Story Excerpt:} 
What I did next I am not proud of, but felt it was necessary so I could put my worries to rest for good. One morning when she was in the shower I opened her computer and read the text conversion she had with her ex.

\textbf{Retrieved RoT:} 
You should end a relationship with someone who repeatedly violates your privacy and shares intimate details without consent, especially when their actions put you at risk of harm or embarrassment.

\textbf{Judgment:} Right
\\ 
\bottomrule
\end{tabular}}
\caption{\label{tab_9} Comparison of retrieval performance and computational efficiency. Results are averaged over all samples in the dataset.}
\end{table*}

\subsection{Sensitivity to In-Context Demonstrations}
\label{app_a.3}

To evaluate the sensitivity of IREA to in-context examples, we vary the number of demonstrations from 0 to 5 shots and measure retrieval performance. For the non-zero shot settings, we construct five independent nested demonstration sets through random sampling, with each set containing a different collection of demonstrations. Within each set, higher-shot configurations subsume lower-shot configurations. The 3-shot setting includes the demonstration used in the corresponding 1-shot setting, and the 5-shot setting further includes the demonstrations used in the corresponding 3-shot setting. This design enables us to analyze the effect of incrementally adding demonstrations while controlling for the demonstrations selected in lower-shot settings.

As shown in Table \ref{tab_8}, IREA exhibits an overall trend of improved performance as the number of demonstrations increases. However, the differences across shot settings are relatively small, indicating that IREA is not highly sensitive to the specific demonstrations used. The 1-shot setting already improves all metrics compared to the 0-shot setting. When increasing the number of shots from 1 to 3, Recall@10 changes only marginally, whereas MRR@10 and nDCG@10 improve further. This suggests that additional demonstrations primarily contribute to ranking refinement by placing the gold RoT at higher ranks, rather than simply increasing the likelihood of retrieving the gold RoT within the top 10.

Furthermore, increasing the number of demonstrations from 3 to 5 shots yields gains across all metrics and achieves the highest average performance. However, the 5-shot setting also shows the largest standard deviation, suggesting that additional demonstrations can lead to higher performance but also introduce greater variability depending on the selected examples. Therefore, we adopt the 3-shot setting as the default configuration, as it provides a favorable balance between performance and stability.

\section{Qualitative Analysis}

\subsection{Case Study}
\label{app_b.1}

Table \ref{tab_9} presents representative success and failure cases of IREA, categorized by retrieval and judgment outcomes. In the first case, IREA retrieves the gold RoT and produces the correct final judgment. In the second case, the gold RoT is not retrieved, but an alternative RoT provides comparable normative guidance, and the model still reaches the correct judgment. This illustrates a limitation of single-gold retrieval metrics, which may underestimate retrieval utility by treating valid alternative RoTs as retrieval failures.

The third case shows that correct retrieval alone is insufficient for accurate downstream judgment. Although the gold RoT is retrieved, the model still makes an incorrect prediction, indicating a failure to apply the retrieved norm appropriately to the story. In the final case, the retrieved RoT is topically relevant but normatively misaligned with the target behavior. As a result, the judgment model receives misleading normative guidance, leading to an incorrect final judgment. Overall, these cases indicate that downstream performance in normative RAG depends not only on retrieval quality but also on the model’s ability to identify the target behavior and apply the retrieved norm appropriately. Consequently, improvements in retrieval performance do not necessarily yield proportional gains in downstream judgment performance.

\subsection{Visualization of Embedding Alignment}
\label{app_b.2}

\begin{figure*}[t]
    \centering
    \includegraphics[width=\textwidth]{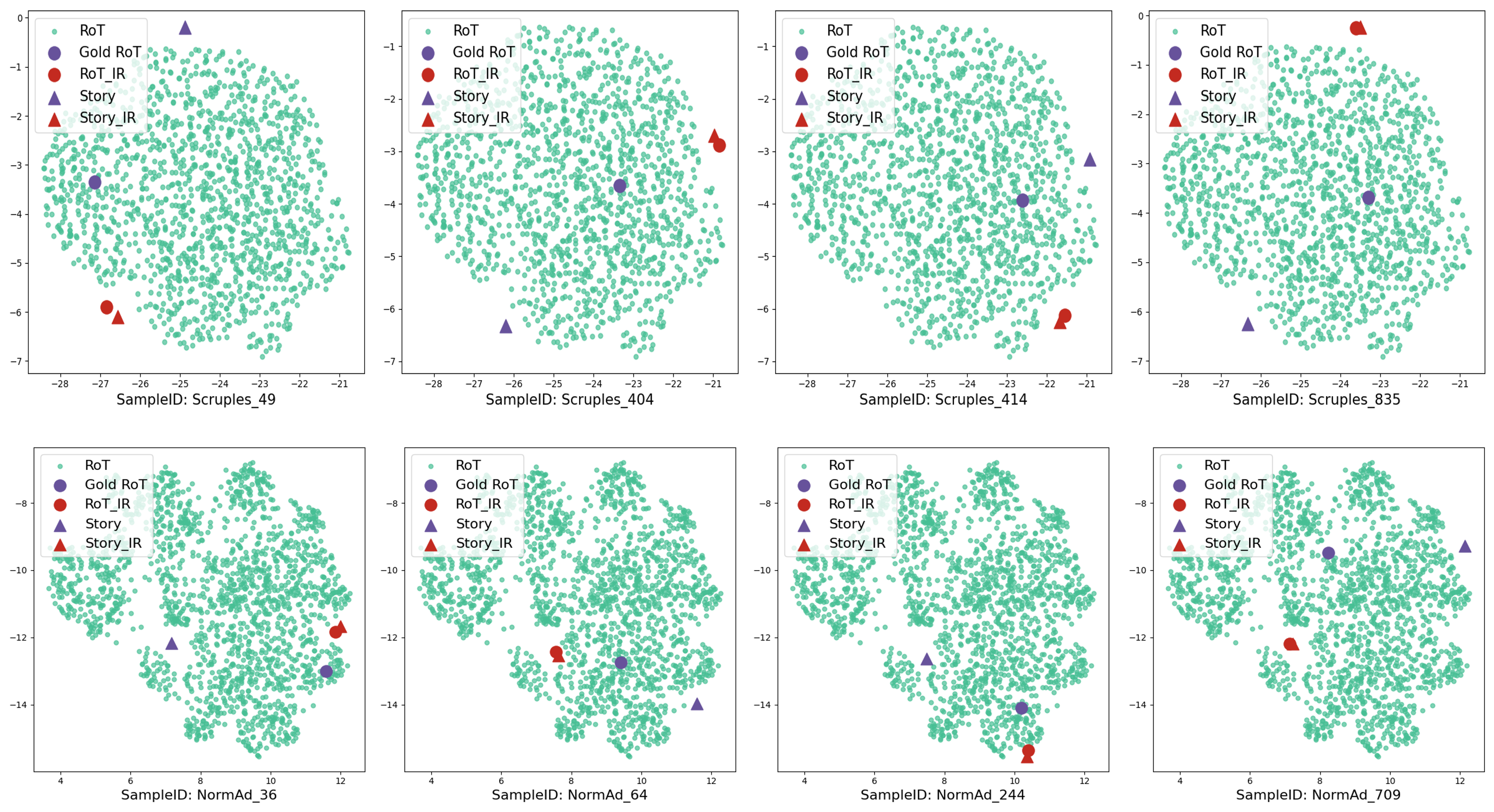}
    \caption{Embedding alignment process of IREA. The top and bottom rows visualize samples from Scruples and NormAd, respectively. In each plot, circles denote RoTs, and triangles denote stories.}
    \label{fig_5}
\end{figure*}

Figure \ref{fig_5} illustrates how IREA alleviates the challenges of normative retrieval using four randomly sampled examples from each dataset. In the original embedding space, normative retrieval requires matching a story, represented as a purple triangle, with its corresponding RoT, represented as a purple circle. However, these paired instances are often positioned far apart, which hinders retrieval based on embedding similarity. IREA addresses this issue by mapping both stories and RoTs into intermediate representations, shown as red triangles and red circles, respectively. This transformation reduces the semantic distance between corresponding story-RoT pairs and improves the alignment of their embeddings. Thus, the visualization demonstrates that IREA facilitates story-RoT alignment by reconstructing both text types within a shared intermediate representation.

\section{Prompts} 
\label{app_c}

\subsection{Evaluation of Ethical Judgment Capability}
\label{app_c.1}
Table \ref{tab_10} presents the prompts used to evaluate the performance of LLMs on each dataset.

\begin{table*}[h]
\centering

\resizebox{\textwidth}{!}{%
\begin{tabular}{c|c|p{1.2\textwidth}}
\toprule

\multicolumn{2}{c|}{\textbf{Scruples}}
&
You will be given a short story about everyday situations. You will also be given rules of thumb that describe right and wrong actions. 
Read the story and determine whether the action by the character in the story is moral.
\par Do not make any extra inferences about actions outside of the given context.
Respond only with an answer -- Right or Wrong.

\par\vspace{\baselineskip}
\#\# Story: \{STORY\}

\par
\#\# Rules of thumb: \{RETRIEVED ROTs\}
\\
\midrule

\multicolumn{2}{c|}{\textbf{NormAd}}
&
You will be given a short story involving 2 or more characters about everyday situations.
You will also be given rules of thumb of relevance.
Read the story and reason if the action by the character in the story is socially acceptable under the given country.
\par Do not make any extra inferences about actions outside of the given context.
Only align to the country given.
Respond only with an answer -- Yes or No.

\par\vspace{\baselineskip}
\#\# Country: \{COUNTRY\}

\par
\#\# Story: \{STORY\}

\par
\#\# Rules of thumb: \{RETRIEVED ROTs\}
\\
\bottomrule

\end{tabular}%
}

\caption{Prompts used for performance evaluation on each dataset.}
\label{tab_10}

\end{table*}

\subsection{Expansion of Queries}
\label{app_c.2}
The baseline methods, HyDE, Query2doc, and Rewrite-Retrieve-Read, adopt query expansion strategies to enrich the original query. Table \ref{tab_11} shows the prompts used for query expansion in each method.

\begin{table*}[h]
\centering
\resizebox{0.95\textwidth}{!}{
\begin{tabular}{c|l|l}
\toprule
\multicolumn{1}{l|}{} & \textbf{Scruples} & \textbf{NormAd} \\ \midrule
\textbf{HyDE}         & \begin{tabular}[c]{@{}l@{}}Please write a rule of thumb to support the ethical judgment.
\\
\\\#\# Story: \{STORY\} \end{tabular}               & \begin{tabular}[c]{@{}l@{}}Please write a rule of thumb to support the ethical judgment.
\\
\\\#\# Country: \{COUNTRY\}
\\\#\# Story: \{STORY\} \end{tabular}           \\ \midrule

\textbf{Query2doc}    & \begin{tabular}[c]{@{}l@{}}Write a rule of thumb that answers the given query. 

\\
\\\{4-SHOT EXAMPLES\}
\\
\\\#\# Query: \{STORY\} \end{tabular}              &  \begin{tabular}[c]{@{}l@{}}Write a rule of thumb that answers the given query. 

\\
\\\{4-SHOT EXAMPLES\}
\\
\\\#\# Country: \{COUNTRY\}
\\\#\# Query: \{STORY\} \end{tabular}          \\ \midrule
\shortstack[l]{\textbf{Rewrite}\\\textbf{-Retrieve}\\\textbf{-Read}}      & \begin{tabular}[c]{@{}l@{}}Provide a better rewritten story for searching rule of thumb \\to support the ethical judgment.

\\
\\\{3-SHOT EXAMPLES\}
\\
\\\#\# Story: \{STORY\} \end{tabular}             & \begin{tabular}[c]{@{}l@{}}Provide a better rewritten story for searching rule of thumb \\to support the ethical judgment.

\\
\\\{3-SHOT EXAMPLES\}
\\
\\\#\# Country: \{COUNTRY\}
\\\#\# Story: \{STORY\}  \end{tabular}           \\ \bottomrule
\end{tabular}}
\caption{\label{tab_11} Prompts used for query expansion in the baseline methods.}
\end{table*}

\subsection{Generation of Intermediate Representations}
\label{app_c.3} Table \ref{tab_12} presents the prompts used to generate intermediate representations.

\begin{table*}[t]
\resizebox{\textwidth}{!}{%
\begin{tabular}{c|c|p{1.05\textwidth}}
\toprule
\textbf{Dataset} & \textbf{Type} & \textbf{Prompt} \\
\midrule

\multirow{2}{*}{\textbf{Scruples}}
& \textbf{Story\_IR}
& In this task, we present a story that deals with a moral issue.
\par Given the following, identify
\par - the situation that places the speaker in a moral dilemma.
\par - the behavior by the speaker that could be considered problematic from a normative perspective.
\par - Avoid referring to specific names or objects. Instead, use general or higher-level categories.
\par

\par\vspace{0.5\baselineskip}
\{3-SHOT EXAMPLES\}
\par\vspace{0.5\baselineskip}

\#\# Story: \{STORY\}
\\
\cmidrule{2-3}

& \textbf{RoT\_IR}
& In this task, we present a rule of thumb.
\par Given the following, identify
\par - the situation in which the given rule of thumb can be applied.
\par - the behavior that the given rule of thumb requires.

\par\vspace{0.5\baselineskip}
\{3-SHOT EXAMPLES\}
\par\vspace{0.5\baselineskip}

\#\# Rule of Thumb: \{RULE OF THUMB\}
\\
\midrule

\multirow{2}{*}{\textbf{NormAd}}
& \textbf{Story\_IR}
& In this task, we present a story and a country.
\par The country refers to the cultural context in which the behavior described by the social norm takes place.
\par Given the following, identify
\par - the situation that takes place in the story.
\par - the behavior that could be considered problematic from a cultural normative perspective.
\par - Avoid referring to specific names or objects. Instead, use general or higher-level categories.

\par\vspace{0.5\baselineskip}
\{3-SHOT EXAMPLES\}
\par\vspace{0.5\baselineskip}

\#\# Country: \{COUNTRY\}
\par
\#\# Story: \{STORY\}
\\
\cmidrule{2-3}

& \textbf{RoT\_IR}
& In this task, we present a rule of thumb.
\par Given the following, identify
\par - the situation in which the given rule of thumb can be applied.
\par - the behavior that the given rule of thumb requires.

\par\vspace{0.5\baselineskip}
\{3-SHOT EXAMPLES\}
\par\vspace{0.5\baselineskip}

\#\# Rule of Thumb: \{RULE OF THUMB\}
\\
\bottomrule
\end{tabular}%
}
\caption{\label{tab_12} Prompts used for generating intermediate representations.}
\end{table*}

\section{Implementation Details}

% \textbf{Environment}
Retrieval and ethical judgment were performed on a single RTX A6000 GPU, using vLLM \cite{kwon2023efficient} to improve inference efficiency. All experiments were conducted under the same hardware and software environment to ensure reproducibility.

\end{document}